# Dictionary Learning and Sparse Coding for Third-order Super-symmetric Tensors

**Piotr Koniusz**[1]     **Anoop Cherian**[2]
[1]LEAR Project Team, Inria Grenoble, France
[2]Australian Centre for Robotic Vision, Australian National University, Canberra
(firstname.lastname)@{inria.fr, anu.edu.au}

## Abstract

Super-symmetric tensors – a higher-order extension of scatter matrices – are becoming increasingly popular in machine learning and computer vision for modeling data statistics, co-occurrences, or even as visual descriptors [15]. However, the size of these tensors are exponential in the data dimensionality, which is a significant concern. In this paper, we study third-order super-symmetric tensor descriptors in the context of dictionary learning and sparse coding. Our goal is to approximate these tensors as sparse conic combinations of atoms from a learned dictionary, where each atom is a symmetric positive semi-definite matrix. Apart from the significant benefits to tensor compression that this framework provides, our experiments demonstrate that the sparse coefficients produced by the scheme lead to better aggregation of high-dimensional data, and showcases superior performance on two common computer vision tasks compared to the state-of-the-art.

## 1 Introduction

Recent times have seen an increasing trend in several machine learning and computer vision applications to use rich data representations that capture the inherent structure or statistics of data. A few notable such representations are histograms, strings, covariances, trees, and graphs. The goal of this paper is to study a new class of structured data descriptors – third-order super-symmetric tensors – in the context of sparse coding and dictionary learning.

Tensors are often used to capture higher-order moments of data distributions such as the covariance, skewness or kurtosis and have been used as data descriptors in several computer vision applications. In region covariances [31], a covariance matrix – computed over multi-modal features from image regions – is used as descriptors for the regions and is seen to demonstrate superior performance in object tracking, retrieval, and video understanding. Given bag-of-words histograms of data vectors from an image, a second-order co-occurrence pooling of these histograms captures the occurrences of two features together in an image and is recently shown to provide superior performance to problems such as semantic segmentation of images, compared to their first-order counterparts [5]. A natural extension of the idea is to use higher-order pooling operators, an extensive experimental analysis of which is provided in [15]. Their paper shows that pooling using third-order super-symmetric tensors can significantly improve upon the second-order descriptors, e.g., by more than 5% MAP on the challenging PASCAL VOC07 dataset.

However, given that the size of the tensors increases exponentially against the dimensionality of their first-order counterpart, efficient representations are extremely important for applications that use these higher-order descriptors. To this end, the goal of this paper is to study these descriptors in the classical dictionary learning and sparse coding setting [23]. Using the properties of super-symmetric tensors, we formulate a novel optimization objective (Sections 4, 5) in which each third-order data tensor is approximated by a sparse non-negative linear combination of positive semi-definite ma-trices. Although our objective is non-convex – typical to several dictionary learning algorithms – we show that our objective is convex in each variable, thus allowing a block-coordinate descent scheme for the optimization. Experiments (presented in Section 6) on the PASCAL VOC07 dataset show that the compressed coefficient vectors produced by our sparse coding scheme preserves the discriminative properties of the original tensors and leads to performances competitive to the state-of-the-art, while providing superior tensor compression and aggregation. Inspired by the merits of third-order pooling techniques proposed in [15], we further introduce a novel tensor descriptor for texture recognition via the linearization of explicitly defined RBF kernels, and show that sparse coding of these novel tensors using our proposed framework provides better performance than the state of the art descriptors used for texture recognition.

### 1.1 Related Work

Dictionary learning and sparse coding [8, 23] methods have significantly contributed to improving the performance of numerous applications in computer vision and image processing. While these algorithms assume a Euclidean vectorial representation of the data, there have been extensions to other data descriptors such as tensors, especially symmetric positive definite matrices [6, 9, 29]. These extensions typically use non-Euclidean geometries defined by a similarity metric that they use for comparing the second-order tensors. Popular choices for comparisons of positive definite matrices are the log-determinant divergence [29], the log-euclidean metric [1], and the affine invariant Riemannian metric [24]. Alas, the third-order tensors considered in this paper are neither positive definite nor there are any standard similarity metrics known apart from the Euclidean distance. Thus, extending these prior methods to our setting is infeasible, demanding novel formulations.

Third-order tensors have been used for various tasks. Spatio-temporal third-order tensor on video data for activity analysis and gesture recognition is proposed in [12] . Non-negative factorization is applied to third-order tensors in [28] for image denoising. Multi-linear algebraic methods for tensor subspace learning are surveyed in [20]. Tensor textures are used in the context of texture rendering in computer vision applications in [33]. Similar to eigenfaces for face recognition, multi-linear algebra based techniques for face recognition use third-order tensors in [32]. However, while these applications generally work with a single tensor, the objective of this paper is to learn the underlying structure of a large collection of such tensors generated from visual data using the framework of dictionary learning and sparse coding, which to the best of our knowledge is a novel proposition.

In addition to the dictionary learning framework, we also introduce an image descriptor for textures. While we are not aware of any previous works that propose third-order tensors as image region descriptors, the most similar methods to our approach are i) third-order probability matrices for image splicing detection [35] and ii) third-order global image descriptors which aggregate SIFT into the autocorrelation tensor [15]. In contrast, our descriptor, is assembled from elementary signals such as luminance and its first- and second-order derivatives, analogous to covariance descriptors [31], but demonstrating superior accuracy. The formulation chosen by us for texture recognition also differs from prior works such as kernel descriptors [2] and convolutional kernel networks [21].

## 2 Notations

Before we proceed, we will briefly review our notations next. We use bold-face upper-case letters for third-order tensors, bold-face lower-case letters for vectors, and normal fonts for scalars. Each second-order tensor along the third mode of a third-order tensor is called a *slice*. Using Matlab style notation, the $s$-th slice of $\boldsymbol{\mathcal{X}}$ is given by $\boldsymbol{\mathcal{X}}_{:,:,s}$. The operation $\uparrow\otimes$ stands for an outer product of a second-order tensor with a vector. For example, $\boldsymbol{\mathcal{Y}} = \mathbf{Y} \uparrow\otimes \mathbf{y}$ produces a third-order tensor $\boldsymbol{\mathcal{Y}} \in \mathbb{R}^{d_1 \times d_2 \times d_3}$ from a matrix $\mathbf{Y} \in \mathbb{R}^{d_1 \times d_2}$ and a vector $\mathbf{y} \in \mathbb{R}^{d_3}$, where the $s$-th slice of $\boldsymbol{\mathcal{Y}}$ is given by $\mathbf{Y}y_s$, $y_s$ being the $s$-th dimension of $\mathbf{y}$. Let $\mathcal{I}_N$ stand for an index set of the first $N$ integers. We use $\mathcal{S}_+^d$ to denote the space of $d \times d$ positive semi-definite matrices and $\mathfrak{S}^d$ to denote the space of super-symmetric tensors of dimensions $d \times d \times d$.

Going by the standard terminology in higher-order tensor factorization literature [7], we define a *core tensor* as the analogous of the singular value matrix in the second-order case. However, unless some specific decompositions are enforced, a core tensor need not be diagonal as in the second-order case.

To this end, we assume the CP decomposition [16] in this paper, which generates a diagonal core tensor.

## 3 Background

In this section, we will review super-symmetric tensors and their properties, followed by a brief exposition of the method described in [15] for generating tensor descriptors for an image. The latter will come useful when introducing our new texture descriptors.

### 3.1 Third-order Super-symmetric Tensors

We define a super-symmetric tensor descriptor as follows:

**Definition 1.** *Suppose $\mathbf{x}_n \in \mathbb{R}_+^d, \forall n \in \mathcal{I}_N$ represents data vectors from an image, then a third-order super-symmetric tensor (TOSST) descriptor $\boldsymbol{\mathcal{X}} \in \mathbb{R}^{d \times d \times d}$ of these data vectors is given by:*

$$\boldsymbol{\mathcal{X}} = \frac{1}{N} \sum_{n=1}^{N} (\mathbf{x}_n \mathbf{x}_n^T) \uparrow \otimes \mathbf{x}_n. \tag{1}$$

In object recognition setting, the data vectors are typically SIFT descriptors extracted from the image, while the tensor descriptor is obtained as the third-order autocorrelation tensor via applying (1).

The following properties of the TOSST descriptor are useful in the sequel.

**Proposition 1.** *For a TOSST descriptor $\boldsymbol{\mathcal{X}} \in \mathbb{R}^{d \times d \times d}$, we have:*

1. *Super-Symmetry: $\boldsymbol{\mathcal{X}}_{i,j,k} = \boldsymbol{\mathcal{X}}_{\Pi(i,j,k)}$ for indexes $i, j, k$ and their permutation given by $\Pi$.*

2. *Every slice is positive semi-definite, that is, $\boldsymbol{\mathcal{X}}_{:,:,s} \in \mathcal{S}_+^d$, for $s \in \mathcal{I}_d$.*

3. *Indefiniteness, i.e., it can have positive, negative, or zero eigenvalues in its core-tensor (assuming a CP decomposition [16]).*

*Proof.* The first two properties can be proved directly from (1). To prove the last property, note that for a TOSST tensor $\boldsymbol{\mathcal{X}}$ and some $\mathbf{z} \in \mathbb{R}^d, \mathbf{z} \neq 0$, we have $((\boldsymbol{\mathcal{X}} \otimes_1 \mathbf{z}) \otimes_2 \mathbf{z}) \otimes_3 \mathbf{z} = \sum_{s=1}^{d} z_s (\mathbf{z}^T \mathbf{X}_s \mathbf{z})$ where $\mathbf{X}_s$ is the $s$-th slice of $\boldsymbol{\mathcal{X}}$. While $\mathbf{z}^T \mathbf{X}_s \mathbf{z} \geq 0$, the tensor product could be negative for $\mathbf{z} < 0$. In the above, $\otimes_i$ denotes tensor product in the $i$-th mode [13]. $\square$

Due to the indefiniteness of the tensors, we cannot use some of the well-known distances on the manifold of SPD matrices [1, 24]. Thus, we restrict to using the Euclidean distance for the derivations and experiments in this paper.

Among several properties of tensors, one that is important and is typically preserved by tensor decompositions is the *tensor rank* defined as:

**Definition 2** (Tensor Rank). *Given a tensor $\boldsymbol{\mathcal{X}} \in \mathbb{R}^{d \times d \times d}$, its tensor rank $\mathrm{TRank}(\boldsymbol{\mathcal{X}})$ is defined as the minimum $p$ such that $\boldsymbol{\mathcal{X}}_{:,:,s} \in \mathrm{Span}(M_1, M_2, ..., M_p)$ for all $s \in \mathcal{I}_d$, where $M_i \in \mathcal{S}_+^d$ are rank-one.*

## 4 Problem Formulation

Suppose we are given data tensors $\boldsymbol{\mathcal{X}}_n, n \in \mathcal{I}_N$, each $\boldsymbol{\mathcal{X}}_n \in \mathfrak{S}^d$. We want to learn a tensor dictionary $\overline{\boldsymbol{\mathcal{B}}}$ with atoms $\boldsymbol{\mathcal{B}}_1, \boldsymbol{\mathcal{B}}_2, ..., \boldsymbol{\mathcal{B}}_K$, where each $\boldsymbol{\mathcal{B}}_k \in \mathfrak{S}^d$ consists of d-slices. Let the $s$-th slice of the $k$-th atom is given by $\mathbf{B}_k^s$ where $s \in \mathcal{I}_d$, $k \in \mathcal{I}_K$. Each slice $\mathbf{B}_k^s \in \mathcal{S}_+^d$. Then, the problem of dictionary learning and sparse coding can be formulated as follows: for sparse coefficient vectors $\boldsymbol{\alpha}^n \in \mathbb{R}^K, n \in \mathcal{I}_N$,

$$\operatorname*{arg\,min}_{\substack{\overline{\boldsymbol{\mathcal{B}}} \\ \boldsymbol{\alpha}^1, ..., \boldsymbol{\alpha}^N}} \sum_{n=1}^{N} \left\| \boldsymbol{\mathcal{X}}_n - \sum_{k=1}^{K} \boldsymbol{\mathcal{B}}_k \alpha_k^n \right\|_F^2 + \lambda \left\| \boldsymbol{\alpha}^n \right\|_1. \tag{2}$$

Note that in the above formulation, third-order dictionary atoms $\mathcal{B}_k$ are multiplied by single scalars $\alpha_k^n$. For $K$ atoms, each of size $d^3$, there are $Kd^3$ parameters to be learned in the dictionary learning process. An obvious consequence of this reconstruction process is that we require $N \gg K$ to prevent overfitting. To circumvent this bottleneck and work in regime $NS \gg K$, the reconstruction process is amended such that every dictionary atom is represented by an outer product of a second-order symmetric positive semi-definite matrix $\mathbf{B}_k$ and a vector $\mathbf{b}_k$. Such a decomposition/approximation will reduce the number of parameters from $Kd^3$ to $K(d^2+d)$. Using this strategy, we re-define dictionary $\overline{\mathcal{B}}$ to be represented by atom pairs $\overline{\mathcal{B}} \equiv \{(\mathbf{B}_k, \mathbf{b}_k)\}_{k=1}^K$ such that $\mathcal{B}_k = \mathbf{B}_k \uparrow \otimes \mathbf{b}_k$. Note that the tensor rank of $\mathcal{B}_k$ under this new representation is equal to the $\text{Rank}(\mathbf{B}_k)$ as formally stated below.

**Proposition 2** (Atom Rank). *Let $\mathcal{B} = \mathbf{B}\uparrow\otimes \mathbf{b}$ and $\|\mathbf{b}\|_1 \neq 0$, then $\text{TRank}(\mathcal{B}) = \text{Rank}(\mathbf{B})$.*

*Proof.* Rank of $\mathcal{B}$ is the smallest $p$ satisfying $(\mathbf{B}\uparrow\otimes\mathbf{b})_{:,:,s} = \mathbf{B} \cdot b_s \in \text{Span}(M_1, M_2, ..., M_p)$, $\forall s = 1, ..., d$, where $M_i \in \mathcal{S}_+^d$ are rank-one. The smallest $p$ satisfying $\mathbf{B} \in \text{Span}(M_1, M_2, ..., M_p)$ is equivalent as multiplication of matrix by any non-zero scalar ($\mathbf{B} \cdot b_s$, $b_s \neq 0$) does not change eigenvectors of this matrix which compose the spanning set. $\square$

Using this idea, we can rewrite (2) as follows:

$$\underset{\substack{\overline{\mathcal{B}} \\ \boldsymbol{\alpha}^1,...,\boldsymbol{\alpha}^N}}{\arg\min} \sum_{n=1}^N \left\| \mathcal{X}_n - \sum_{k=1}^K (\mathbf{B}_k \uparrow \otimes \mathbf{b}_k) \alpha_k^n \right\|_F^2 + \lambda \|\boldsymbol{\alpha}^n\|_1. \quad (3)$$

Introducing optimization variables $\boldsymbol{\beta}_k^n$ and rewriting the loss function by taking the slices out, we can further rewrite (3) as:

$$\underset{\substack{\overline{\mathcal{B}} \\ \boldsymbol{\alpha}^1,...,\boldsymbol{\alpha}^N}}{\arg\min} \sum_{n=1}^N \sum_{s=1}^S \left\| \mathbf{X}_n^s - \sum_{k=1}^K \mathbf{B}_k \beta_k^{s,n} \right\|_F^2 + \lambda \|\boldsymbol{\alpha}^n\|_1,$$

$$\text{subject to } \boldsymbol{\beta}_k^n = \mathbf{b}_k \alpha_k^n, \quad \forall k \in \mathcal{I}_K \text{ and } n \in \mathcal{I}_N. \quad (4)$$

We may rewrite the equality constraints in (4) as a proximity constraint in the objective function (using a regularization parameter $\gamma$), and constrain the sparse coefficients in $\boldsymbol{\alpha}_n$ to be non-negative, as each slice is positive semi-definite. This is a standard technique used in optimization, commonly referred to as *variable splitting*. We can also normalize the dictionary atoms $\mathbf{B}_k$ to be positive semi-definite and of unit-trace for better numerical stability. In that case, vector atoms $\mathbf{b}_k$ can be constrained to be non-negative (or even better, $\boldsymbol{\beta}_k^n \geq 0$ instead of $\boldsymbol{\alpha}_n \geq 0$ and $\mathbf{b}_k \geq 0$). Introducing chosen constraints, we can rewrite our tensor dictionary learning and sparse coding formulation:

$$\underset{\substack{\overline{\mathcal{B}} \\ \boldsymbol{\alpha}^1,...,\boldsymbol{\alpha}^N \geq 0}}{\arg\min} \sum_{n=1}^N \sum_{s=1}^S \left\| \mathbf{X}_n^s - \sum_{k=1}^K \mathbf{B}_k \beta_k^{s,n} \right\|_F^2 + \gamma \|\boldsymbol{\beta}^{s,n} - \mathbf{b}^s \odot \boldsymbol{\alpha}^n\|_2^2 + \lambda \|\boldsymbol{\alpha}^n\|_1,$$

$$\text{subject to } \mathbf{B}_k \succcurlyeq 0, \ \text{Tr}(\mathbf{B}_k) \leq 1, \ \|\mathbf{b}_k\|_1 \leq 1, \quad k=1,...,K. \quad (5)$$

In the above equation, operator $\odot$ is an element-wise multiplication between vectors. Coefficients $\gamma$ and $\lambda$ control the quality of the proximity and sparsity of $\boldsymbol{\alpha}_n$, respectively. We might replace the loss function using a loss based on Riemannian geometry of positive definite matrices [6], in which case the convexity would be lost and thus convergence cannot be guaranteed. The formulation (5) is convex in each variable and can be solved efficiently using the block coordinate descend.

**Remark 1** (Non-symmetric Tensors). *Note that non-symmetric $\mathcal{X}_n \in \mathbb{R}^{d_1 \times d_2 \times d_3}$ can be coded if the positive semi-definite constraint on $\mathbf{B}_k$ is dropped, $\mathbf{B}_k \in \mathbb{R}^{d_1 \times d_2}$ and $\mathbf{b}_k \in \mathbb{R}^{d_3}$. Other non-negative constraints can also be then removed. While this case is a straightforward extension of our formulation, a thorough investigation into it is left as future work.*

**Optimization**

We propose a block-coordinate descent scheme to solve (5), in which each variable in the formulation is solved separately, while keeping the other variables fixed. As each sub-problem is convex, we are guaranteed to converge to a local minimum. Initial values of $\mathbf{B}_k$, $\mathbf{b}_k$, and $\boldsymbol{\alpha}^n$ are chosen randomly from the uniform distribution within a prespecified range. Vectors $\boldsymbol{\beta}^{s,n}$ are initialized with the Lasso algorithm. Next, we solve four separate convex sub-problems resulting in updates of $\mathbf{B}_k$, $\mathbf{b}_k$, $\boldsymbol{\beta}^{s,n}$, $\boldsymbol{\alpha}^n$. For learning second-order matrices $\mathbf{B}_k$, we employ the PQN solver [27] which enables us to implement and apply projections into the SPD cone, handle the trace norm or even the rank of matrices (if this constraint is used). Similarly, the projected gradient is used to keep $\mathbf{b}_k$ inside the simplex $\|\mathbf{b}_k\|_1 \leq 1$. For the remaining sub-problems, we use the L-BFGS-B solver [4] which enables us to handle simple box constraints e.g., we split coding for $\boldsymbol{\alpha}^n$ into two non-negative sub-problems $\boldsymbol{\alpha}^n_+$ and $\boldsymbol{\alpha}^n_-$ by applying box constraints $\boldsymbol{\alpha}^n_+ \geq 0$ and $\boldsymbol{\alpha}^n_- \geq 0$. Finally, we obtain $\boldsymbol{\alpha}^n = \boldsymbol{\alpha}^n_+ - \boldsymbol{\alpha}^n_-$ after minimization. Moreover, for the coding phase (no dictionary learning), we fix $\{(\mathbf{B}_k, \mathbf{b}_k)\}_{k=1}^K$. The algorithm proceeds as above, but without updates of $\mathbf{B}_k$ and $\mathbf{b}_k$. Algorithm 1 shows the important steps of our algorithm; this includes both the dictionary learning and the sparse coding sub-parts.

---

**Algorithm 1:** Third-order Sparse Dictionary Learning and Coding.

**Data**: $N$ data tensors $\overline{\boldsymbol{\mathcal{X}}} \equiv \{\boldsymbol{\mathcal{X}}_1, ..., \boldsymbol{\mathcal{X}}_N\}$, proximity and sparsity constants $\gamma$ and $\lambda$, stepsize $\eta$, $K$ dictionary atoms $\overline{\boldsymbol{\mathcal{B}}} \equiv \{(\mathbf{B}_k, \mathbf{b}_k)\}_{k=1}^K$ if coding, otherwise $LearnDict\!=\!true$

**Result**: $N$ sparse coeffs. $\overline{\boldsymbol{\alpha}} \equiv \{\boldsymbol{\alpha}^1, ..., \boldsymbol{\alpha}^N\}$, $K$ atoms $\overline{\boldsymbol{\mathcal{B}}} \equiv \{(\mathbf{B}_k, \mathbf{b}_k)\}_{k=1}^K$ if $LearnDict\!=\!true$

**Initialization:**
**if** $LearnDict$ **then**
- Uniformly sample slices from $\overline{\boldsymbol{\mathcal{X}}}$ and fill $\mathbf{B}_k, \forall k \!\in\! \mathcal{I}_K$
- Uniformly sample values from $-1/K$ to $1/K$ and initialize $\mathbf{b}_k, \forall k \!\in\! \mathcal{I}_K$

**end**
- Vectorize inputs $\mathbf{X}_n^s$ and atoms $\mathbf{B}_k, \forall s \!\in\! \mathcal{I}_S, \forall n \!\in\! \mathcal{I}_N, \forall k \!\in\! \mathcal{I}_K$
  solve Lasso problem and fill $\boldsymbol{\beta}^{s,n}, \forall s \!\in\! \mathcal{I}_S, \forall n \!\in\! \mathcal{I}_N$
- Uniformly sample values from $-1/K\lambda$ to $1/K\lambda$ and fill $\boldsymbol{\alpha}^n, \forall n \!\in\! \mathcal{I}_N$
- $Objective^{(0)}\!=\!0$, $t\!=\!1$

**Main loop:**
**while** $\neg Converged$ **do**
    **if** $LearnDict$ **then**
- $\mathbf{B}_k^{(t+1)} = \Pi_{SPD}\left(\mathbf{B}_k^{(t)} - \eta \left.\frac{\partial f}{\partial \mathbf{B}_k}\right|_{\mathbf{B}_k^{(t)}}\right), \forall k \!\in\! \mathcal{I}_K$, where $f$ is the cost from (5) and
  projection $\Pi_{SPD}(\mathbf{B}) = \frac{\mathbf{B}^*}{\text{Tr}(\mathbf{B}^*)}, \mathbf{B}^* \!=\! \boldsymbol{U} \max(\boldsymbol{\lambda}^*, 0)\boldsymbol{V}^T$ for $(\boldsymbol{U}, \boldsymbol{\lambda}^*, \boldsymbol{V}) \!=\! SVD(\mathbf{B})$
- $\mathbf{b}_k^{(t+1)} = \Pi_+\left(\mathbf{b}_k^{(t)} - \eta \left.\frac{\partial f}{\partial \mathbf{b}_k}\right|_{\mathbf{b}_k^{(t)}}\right), \forall k \!\in\! \mathcal{I}_K$, and $\Pi_+(\mathbf{b}) \!=\! \max(\mathbf{b}, 0)$ if $\Pi_+$ is used

    **end**
- $\boldsymbol{\beta}^{s,n,(t+1)} = \Pi_+\left(\boldsymbol{\beta}^{s,n,(t)} - \eta \left.\frac{\partial f}{\partial \boldsymbol{\beta}}\right|_{\boldsymbol{\beta}_k^{s,n,(t)}}\right), \forall s \!\in\! \mathcal{I}_S, \forall n \!\in\! \mathcal{I}_N$, use of $\Pi_+$ is optional
- $\boldsymbol{\alpha}_n^{(t+1)} = \Pi_+\left(\boldsymbol{\alpha}_n^{(t)} - \eta \left.\frac{\partial f}{\partial \boldsymbol{\alpha}}\right|_{\boldsymbol{\alpha}_n^{(t)}}\right), \forall n \!\in\! \mathcal{I}_N$, use of $\Pi_+$ is optional
- $Objective^{(t)} = f\left(\overline{\boldsymbol{\mathcal{X}}}, \overline{\mathbf{B}}^{(t)}, \overline{\mathbf{b}}^{(t)}, \overline{\boldsymbol{\alpha}}^{(t)}\right)$, where $\overline{\mathbf{B}}^{(t)}$ and $\overline{\mathbf{b}}^{(t)}$ are $K$ matrix and vector atoms
- $Converged = EvaluateStopingCriteria\left(t, Objective^{(t)}, Objective^{(t-1)}\right)$

**end**

---

## 5 Theory

In this section, we analyze some of the theoretical properties of our sparse coding formulation. We establish that under an approximation setting, a sparse coding of the data tensor exist in some dictionary for every data point. We also provide bounds on the quality of this approximation and its tensor rank. The following lemma comes useful in the sequel.

**Lemma 1** ([30]). *Let* $\mathbf{Y} = \sum_{i=1}^{m} \boldsymbol{\phi}_i \boldsymbol{\phi}_i^T$, *where each* $\boldsymbol{\phi}_i \in \mathbb{R}^d$ *and* $m = \Omega(d^2)$. *Then, there exist an* $\boldsymbol{\alpha} \in \mathbb{R}_+^m$ *and* $\epsilon \in (0, 1)$ *such that:*

$$\mathbf{Y} \preceq \sum_{i=1}^{m} \alpha_i \boldsymbol{\phi}_i \boldsymbol{\phi}_i^T \preceq (1 + \epsilon) \mathbf{Y}, \tag{6}$$

*where* $\boldsymbol{\alpha}$ *has* $\mathcal{O}(d \log(d)/\epsilon^2)$ *non-zeros.*

**Theorem 1.** *Let* $\mathcal{X} = \sum_{i=1}^{m} (\boldsymbol{\phi}_i \boldsymbol{\phi}_i^T) \uparrow \otimes \boldsymbol{\phi}_i$, *then there exist a second-order dictionary* $\mathbf{B}$ *with* $md$ *atoms and a sparse coefficient vector* $\boldsymbol{\beta} \in \mathbb{R}_+^{md}$ *with* $\mathcal{O}(d^2 \log(d)/\epsilon^2)$ *non-zeros, such that for* $\epsilon \in (0, 1)$,

$$I) \quad \widetilde{\mathcal{X}} = \sum_{i=1}^{m} (\boldsymbol{\phi}_i \boldsymbol{\phi}_i^T) \uparrow \otimes \bar{\boldsymbol{\beta}}^i \quad and \quad II) \quad \left\| \mathcal{X} - \widetilde{\mathcal{X}} \right\|_F \leq \epsilon \sum_{s=1}^{d} \|\mathbf{X}_s\|_F ,$$

*where* $\mathbf{X}_s$ *is the s-th slice of* $\mathcal{X}$ *and* $\bar{\boldsymbol{\beta}}^i \in \mathbb{R}^d$.

*Proof.* To prove I: each slice $\mathbf{X}_s = \sum_{i=1}^{m} (\boldsymbol{\phi}_i \boldsymbol{\phi}_i^T) \phi_i^s$, where $\phi_i^s$ is the $s$-th dimension of $\boldsymbol{\phi}_i$. Let $\boldsymbol{\phi}_i' = \boldsymbol{\phi}_i \sqrt{\phi_i^s}$, then $\mathbf{X}_s = \sum_{i=1}^{m} \boldsymbol{\phi}_i' \boldsymbol{\phi}_i'^T$. If the slices are to be treated independently (as in (5)), then to each slice we can apply Lemma 1 that results in sparse coefficient vectors $\bar{\boldsymbol{\beta}}^i$ having $\mathcal{O}(d \log(d)/\epsilon^2)$ non-zeros. Extending this result to all the $d$-slices, we obtain the result.
To prove II: substituting $\widetilde{\mathcal{X}}$ with its sparse approximation and using the upper-bound in (6) for each slice, the result follows. □

**Theorem 2** (Approximation quality via Tensor Rank). *Let* $\widetilde{\mathcal{X}}$ *be the sparse approximation to* $\mathcal{X}$ *obtained by solving* (4) *using a dictionary* $\mathbf{B}$ *and if* $\mathrm{Rank}(\mathbf{B})$ *represent the maximum of the rank of any atom, then*

$$\mathrm{TRank}(\widetilde{\mathcal{X}}) \leq \left| \bigcup_{s=1}^{d} \mathrm{Supp}(\mathbf{X}_s) \right| \mathrm{Rank}(\mathbf{B}), \tag{7}$$

*where* $\mathrm{Supp}(\mathbf{X}_s)$ *is the support set produced by* (4) *for slice* $\mathbf{X}_s$.

*Proof.* Rewriting $\widetilde{\mathcal{X}}$ in terms of its sparse tensor approximation, followed by applying Theorem 2 and the definition in (2), the proof directly follows. □

Theorem 2 gives a bound on the rank of approximation of a tensor $\mathcal{X}$ by $\widetilde{\mathcal{X}}$ (which is a linear combination of atoms). Knowing rank of $\mathcal{X}$ and $\widetilde{\mathcal{X}}$ helps check how good the approximation is over measuring the simple Euclidean norm between them. This is useful in experiments where we impose rank-r (r=1,2,...) constraints on atoms $\mathbf{B}_k$ and measures how rank constraints on $\mathbf{B}_k$ impacts the quality of approximation.

## 6 Experiments

In this section, we present experiments demonstrating the usefulness of our framework. As second-order region covariance descriptors (RCD) are the most similar tensor descriptors to our TOSST descriptor, we decided to evaluate our performance on applications of RCDs, specifically for texture recognition. In the sequel, we propose a novel TOSST texture descriptor. This will follow experiments evaluating this new descriptor to state-of-the-art on two texture benchmarks, namely the Brodatz textures [22] and the UIUC materials [18]. Moreover, experiments illustrating behaviour of our dictionary learning are provided. We also demonstrate the adequacy of our framework to compression of third-order global image descriptors [15] on the challenging PASCAL VOC07 set.

## 6.1 TOSST Texture Descriptors

Many region descriptors for the texture recognition are described in the literature [5, 17, 31]. Their construction generally consists of the following steps:

i) For an image $\mathbf{I}$ and a region in it $\mathcal{R}$, extract feature statistics from the region: if $(x_i, y_i)$ represent pixel coordinates in $\mathcal{R}$, then, a feature vector from $\mathcal{R}$ will be:

$$\phi_{xy}(\mathbf{I}) = \left[\frac{x - x_0}{w - 1}, \frac{y - y_0}{h - 1}, I_{xy}, \frac{\partial I_{xy}}{\partial x}, \frac{\partial I_{xy}}{\partial y}, \frac{\partial^2 I_{xy}}{\partial x^2}, \frac{\partial^2 I_{xy}}{\partial y^2}\right]^T, \quad (8)$$

where $(x_0, y_0)$ and $(w, h)$ are the coordinate origin, width, and height of $\mathcal{R}$, respectively.

ii) Compute a covariance matrix $\boldsymbol{\Phi}(\mathbf{I}, \mathcal{R}) = \frac{1}{|\mathcal{R}|-1} \sum_{(x,y)\in\mathcal{R}} \left(\phi_{xy}(\mathbf{I}) - \bar{\phi}\right)\left(\phi_{xy}(\mathbf{I}) - \bar{\phi}\right)^T$, where $\bar{\phi}$ is the mean over $\phi_{xy}(\mathbf{I})$ from that region [31, 17]. Alternatively, one can simply compute an autocorrelation matrix $\boldsymbol{\Phi}(\mathbf{I}, \mathcal{R}) = \frac{1}{|\mathcal{R}|} \sum_{(x,y)\in\mathcal{R}} \phi_{xy}(\mathbf{I})\phi_{xy}(\mathbf{I})^T$ as described in [5, 15].

In this work, we make modifications to the steps listed above to work with RBF kernels which are known to improve the classification performance of descriptor representations over linear methods. The concatenated statistics in step (i) can be seen as linear features forming linear kernel $K_{lin}$:

$$K_{lin}((x, y, \mathbf{I}^a), (x', y', \mathbf{I}^b)) = \phi_{xy}(\mathbf{I}^a)^T \phi_{x'y'}(\mathbf{I}^b). \quad (9)$$

We go a step further and define the following sum kernel $K_{rbf}$ composed of several RBF kernels $G$:

$$K_{rbf}\left((x, y, \mathbf{I}^a), (x', y', \mathbf{I}^b)\right) = \sum_{i=1}^{7} G_{\sigma_i}^i(\phi_{xy}^i(\mathbf{I}^a) - \phi_{x'y'}^i(\mathbf{I}^b)), \quad (10)$$

where $\phi_{xy}^i(\mathbf{I})$ is the $i$-th feature in (8) and $G_{\sigma_i}^i(\mathbf{u} - \mathbf{u}') = \exp(-\left\|\mathbf{u} - \mathbf{u}'\right\|_2^2 / 2\sigma_i^2)$, represents so-called relative compatibility kernel that measures the compatibility between the features in each of the image regions.

The next step involves linearization of Gaussian kernels $G_\sigma^i$ for $i = 1, ..., \tau$ to obtain feature maps that expresses the kernel $K_{rbf}$ by the dot-product. In what follows, we use a fast approximation method based on probability product kernels [10]. Specifically, we employ the inner product for the $d'$-dimensional isotropic Gaussian distribution:

$$G_\sigma\left(\mathbf{u} - \mathbf{u}'\right) = \left(\frac{2}{\pi\sigma^2}\right)^{\frac{d'}{2}} \int_{\boldsymbol{\zeta}\in\mathbb{R}^{d'}} G_{\sigma/\sqrt{2}}\left(\mathbf{u} - \boldsymbol{\zeta}\right) G_{\sigma/\sqrt{2}}\left(\mathbf{u}' - \boldsymbol{\zeta}\right) \mathrm{d}\boldsymbol{\zeta}. \quad (11)$$

Equation (11) is then approximated by replacing the integral with the sum over $Z$ pivots $\boldsymbol{\zeta}_1, ..., \boldsymbol{\zeta}_Z$:

$$G_\sigma(\mathbf{u} - \mathbf{u}') \approx \left\langle\sqrt{w}\boldsymbol{\phi}(\mathbf{u}), \sqrt{w}\boldsymbol{\phi}(\mathbf{u}')\right\rangle, \; \boldsymbol{\phi}(\mathbf{u}) = \left[G_{\sigma/\sqrt{2}}(\mathbf{u} - \boldsymbol{\zeta}_1), ..., G_{\sigma/\sqrt{2}}(\mathbf{u} - \boldsymbol{\zeta}_Z)\right]^T. \quad (12)$$

A weighting constant $w$ relates to the width of the rectangle in the numerical integration and is factored out by the $\ell_2$ normalization:

$$G_\sigma(\mathbf{u} - \mathbf{u}') = \frac{G_\sigma(\mathbf{u} - \mathbf{u}')}{G_\sigma(\mathbf{u} - \mathbf{u})G_\sigma(\mathbf{u}' - \mathbf{u}')} \approx \left\langle\frac{\sqrt{w}\boldsymbol{\phi}(\mathbf{u})}{\|\sqrt{w}\boldsymbol{\phi}(\mathbf{u})\|_2}, \frac{\sqrt{w}\boldsymbol{\phi}(\mathbf{u}')}{\|\sqrt{w}\boldsymbol{\phi}(\mathbf{u}')\|_2}\right\rangle = \left\langle\frac{\boldsymbol{\phi}(\mathbf{u})}{\|\boldsymbol{\phi}(\mathbf{u})\|_2}, \frac{\boldsymbol{\phi}(\mathbf{u}')}{\|\boldsymbol{\phi}(\mathbf{u}')\|_2}\right\rangle. \quad (13)$$

The task of linearization of each $G_{\sigma_i}^i$ in (10) becomes trivial as these kernels operate on one-dimensional variables ($d' = 1$). Therefore, we sample uniformly domains of each variable and use $Z = 5$. With this tool at hand, we rewrite kernel $K_{rbf}$ as:

$$K_{rbf}((x, y, \mathbf{I}^a), (x', y', \mathbf{I}^b)) \approx \left\langle\boldsymbol{v}_{xy}^a, \boldsymbol{v}_{x'y'}^b\right\rangle, \quad (14)$$

where vector $\boldsymbol{v}^a_{xy}$ is composed by $\tau$ sub-vectors $\frac{\phi^i_{xy,\sigma_i}}{\left\|\phi^i_{xy,\sigma_i}\right\|_2}$ and each sub-vector $i = 1, ..., \tau$ is a result of linearization by equations (11-13). We use a third-order aggregation of $\boldsymbol{v}$ according to equation (1) to generate our TOSST descriptor for textures. For comparisons, we also aggregate $\phi$ from equation (8) into a third-order linear descriptor. See our supplementary material for the definition of the third-order global image descriptors [15] followed by the derivation of the third-order aggregation procedure based on kernel exponentiation.

### 6.2 Datasets

The Brodatz dataset contains 111 different textures each one represented by a single 640×640 image. We follow the standard protocol, i.e., each texture image is subdivided into 49 overlapping blocks of size 160×160. For training, 24 randomly selected blocks are used per class, while the rest is used for testing. We use 10 data splits. The UIUC materials dataset contains 18 subcategories of materials taken in the wild from four general categories e.g., bark, fabric, construction materials, and outer coat of animals. Each subcategory contains 12 images taken at various scales. We apply a leave-one-out evaluation protocol [18]. The PASCAL VOC07 set consists of 5011 images for training and 4952 for testing, 20 classes of objects of varied nature e.g., human, cat, chair, train, bottle.

### 6.3 Experimental Setup

**TOSST descriptors.** In what follows, we evaluate the following variants of the TOSST descriptor on the Brodatz dataset: linear descriptor with first- and -second order derivatives as in (8, 9) (i) but without luminance, and ii) with luminance. RBF descriptor as in (8, 10) (iii) without luminance, and iv) with luminance. Moreover, v) is a variant based on (iv) that uses the opponent color cues for evaluations on the UIUC materials set [18]. Removing luminance for (i) and (iii) helps emphasize the benefit of RBF over linear formulation. First though, we investigate our dictionary learning.

**Dictionary learning.** We evaluate several variants of our dictionary learning approach on the Brodatz dataset. We use the RBF descriptor without the luminance cue (iii) to prevent saturation in performance. We apply patch size 40 with stride 20 to further lower computational complexity, sacrificing performance slightly. This results in 49 TOSST descriptors per block. We do not apply any whitening, thus preserving the SPD structure of the TOSST slices.

Figures 1(a) and 1(b) plot the accuracy and the dictionary learning objective against an increasing size $K$ of the dictionary. We analyze three different regularization schemes on the second-order dictionary atoms in (5), namely (1) with only SPD constraints, (2) with SPD and low-rank constraints, and (3) without SPD or low-rank constraints. When we enforce $\text{Rank}(\mathbf{B}_k) \leq R < d$ for $k = 1, ..., K$ to obtain low-rank atoms $\boldsymbol{\mathcal{B}}_k$ by scheme (2), the convexity w.r.t. $\mathbf{B}_k$ is lost because the optimization domain in this case is the non-convex boundary of the PD cone. However, the formulation (5) for schemes (1) and (3) is convex w.r.t. $\mathbf{B}_k$. The plots highlight that for lower dictionary sizes, variants (1) and (2) perform worse than (3) which exhibits the best performance due to a very low objective. However, the effects of overfitting start emerging for larger $K$. The *SPD* constraint appears to offer the best trade-off resulting in a good performance for both small and large $K$. In Figure 1(c), we further evaluate the performance for a *Rank-R* constraint given $K = 20$ atoms. The plot highlights that imposing the low-rank constraint on atoms may benefit classification. Note that for $15 \leq R \leq 30$, the classification accuracy fluctuates up to $2\%$, which is beyond the error standard deviation ($\leq 0.4\%$) despite any significant fluctuation to the objective in this range. This suggests that imposing some structural constraints on dictionary atoms is an important step in dictionary learning. Lastly, Figure 1(d) shows the classification accuracy for the coding and pooling steps w.r.t. $\lambda$ controlling sparsity and suggests that the texture classification favors low sparsity.

|  | (i) linear, no lum. | (ii) linear, lum. | (iii) RBF, no lum. | (iv) RBF, lum. | (v) RBF, lum., opp. |
|---|---|---|---|---|---|
| dataset | Brodatz | Brodatz | Brodatz | Brodatz | UIUC materials |
| ten. size $d$ | 6 | 7 | 30 | 35 | 45 |
| accuracy | 93.9±0.2% | 99.4±0.1% | 99.4±0.2% | **99.9±0.08%** | **58.0±4.3%** |

| Brodatz | ELBCM 98.72% [26] | L$^2$ECM 97.9% [17] | RC 97.7% [31] |
|---|---|---|---|
| UIUC materials | SD 43.5% [18] | CDL 52.3±4.3% [34] | RSR 52.8±5.1% [9] |

**Table 1:** Evaluations of the proposed TOSST descriptor and comparisons to the state-of-the-art.

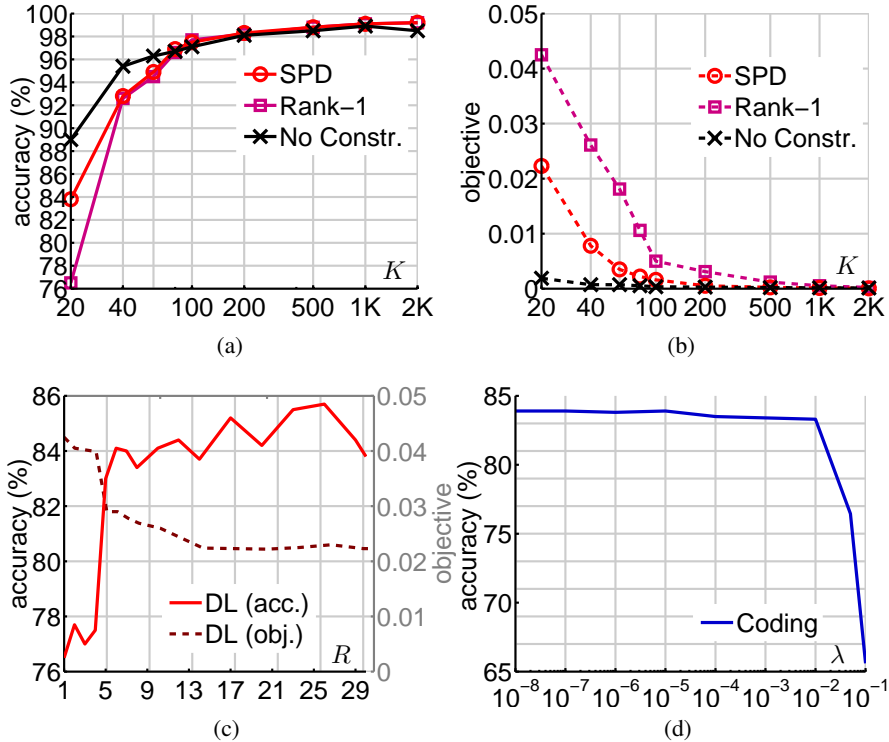

**Figure 1:** Parameter sensitivity on Brodatz textures: 1(a) and 1(b) show classification accuracy and objective values against various dictionary sizes respectively, 1(c) and 1(d) show accuracy with fixed $K = 20$, but varying the atom rank $R$, and sparsity regularization $\lambda$.

### 6.4 Comparison to State-of-the-Art on Textures

In this experiment, descriptor variants were whitened per block as in [15]. In each block, we extract 225 TOSST descriptors with patch size 20 and stride 10, apply our dictionary learning given $K = 2000$ followed by the sparse coding and perform pooling as in [14] prior to classification with SVM. Table 1 demonstrates our results which highlight that the RBF variant outperforms the linear descriptor. This is even more notable when the luminance cue is deliberately removed from descriptors to degrade their expressiveness. Table 1 also demonstrates state-of-the-art results. With $99.9 \pm 0.08\%$ accuracy, our method is the strongest performer. Additionally, Table 1 provides our results on the UIUC materials obtained with descriptor variant (v) (described above) and $K = 4000$ dictionary atoms. We used TOSST descriptors with patch size 40, stride 20 and obtained $58.0 \pm 4.3\%$ which outperforms other methods.

### 6.5 Signature Compression on PASCAL VOC07

In this section, we evaluate our method on the PASCAL VOC07 dataset, which is larger and more challenging than the textures. To this end, we use the setup in [15] to generate third-order global image descriptors (also detailed in our supp. material) with the goal of compressing them using our sparse coding framework. In detail, we use SIFT extracted from gray-scale images with radii $12, 16, 24, 32$ and stride $4, 6, 8, 10$, reduce SIFT size to $90D$, append SPM codes of size $11D$ [15] and obtain the global image signatures which yield baseline score of $61.3\%$ MAP as indicated in Table 2 (last column). Next, we learn dictionaries using our setup as in equation (5) and apply our sparse coding to reduce signature dimensionality from $176851$ (upper simplex of third-order tensor) to sizes from $2K$ to $25K$. The goal is to regain the baseline score. Table 2 shows that a learned dictionary *(DL)* of size $25K$ yields $61.2\%$ MAP at $7\times$ compression rate, which demonstrates how to limit redundancy in third-order tensors. For comparison, a random dictionary is about $4.3\%$ worse. In Figure 2 plots the impact of the number of atoms ($K$) against signature compression on this dataset.

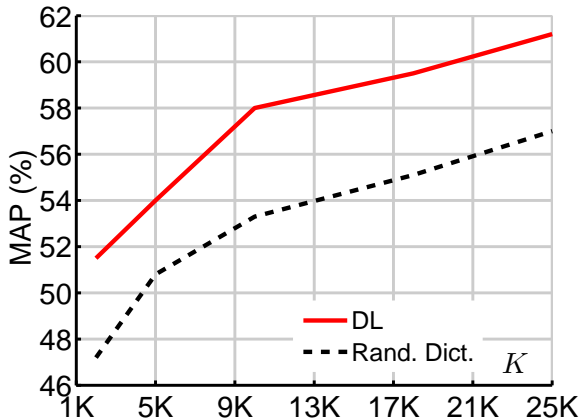

**Figure 2:** This plot illustrates the impact of $K$ on the signature compression on PASCAL VOC07. Dictionary learning (*DL*) performs better than *(Random Dict.)* formed by sampling atoms from a distribution of DL.

| $K$ | 2000 | 5000 | 10000 | 18000 | 25000 | 176851 |
|---|---|---|---|---|---|---|
| DL (MAP%) | 51.5% | 54.0% | 58.0% | 59.5% | **61.2%** | 61.3% |
| Random Dict. (MAP%) | 47.2% | 50.8% | 53.3% | 55.1% | 57.0% | |

**Table 2:** Evaluations of the compression w.r.t. $K$ on third-order global descriptors and PASCAL VOC07.

**Complexity.** It takes about $3.2s$ to code a third-order tensor of size $d = 30$ using a dictionary with 2000 atoms. Note that our current implementation is in MATLAB and the performance is measured on a 4-core CPU. Our dictionary learning formulation converges in approximately 50 iterations.

## 7 Conclusions

We presented a novel formulation and an efficient algorithm for sparse coding third-order tensors using a learned dictionary consisting of second-order positive semi-definite atoms. Our experiments demonstrate that our scheme leads to significant compression of the input descriptors, while not sacrificing the classification performance of the respective application. Further, we proposed a novel tensor descriptor for texture recognition, which when sparse-coded by our approach, achieves state-of-the-art performance on two standard benchmark datasets for this task.

**Acknowledgements.** We thank several anonymous reviewers for their very positive reviews and insightful questions that helped improve this paper.

# Appendices

## A   Third-order Global Image Descriptor [15]

In what follows, we outline a global image descriptor from [15] used in our experiments on the PascalVOC07 dataset. Third-order Global Image Descriptor is based on SIFT [19] aggregated into a third-order autocorrelation tensor followed by Higher Order Singular Value Decomposition [16, 13] used for the purpose of signal whitening. This is achieved by Power Normalisation [15] which is performed on the eigenvalues from so-called core tensor. The following steps from [15] are used:

$$\boldsymbol{\mathcal{V}}_i = \otimes_r \mathbf{v}_i, \ \forall i \in \mathcal{I} \tag{15}$$

$$\bar{\boldsymbol{\mathcal{V}}} = \underset{i \in \mathcal{I}}{\mathrm{Avg}}(\boldsymbol{\mathcal{V}}_i) \tag{16}$$

$$(\boldsymbol{E}; \boldsymbol{A}) = \mathrm{HOSVD}(\bar{\boldsymbol{\mathcal{V}}}) \tag{17}$$

$$\hat{\boldsymbol{E}} = \mathrm{Sgn}(\boldsymbol{E}) \, |\boldsymbol{E}|^{\gamma_e} \tag{18}$$

$$\hat{\boldsymbol{\mathcal{V}}} = \hat{\boldsymbol{E}} \times_1 \boldsymbol{A} ... \times_r \boldsymbol{A} \tag{19}$$

$$\boldsymbol{\mathcal{X}} = \mathrm{Sgn}(\hat{\boldsymbol{\mathcal{V}}}) \, |\hat{\boldsymbol{\mathcal{V}}}|^{\gamma_c} \tag{20}$$

Equations (15, 16) assemble a higher-order autocorrelation tensor $\bar{\boldsymbol{\mathcal{V}}}$ per image $\mathcal{I}$. First, the outer product $\otimes_r$ of order $r$ [16, 13, 15] is applied to the local image descriptors $\mathbf{v}_i \in \mathbb{R}^d$ from $\mathcal{I}$. This results in $|\mathcal{I}|$ autocorrelation matrices or third-order tensors $\boldsymbol{\mathcal{V}}_i \in \mathbb{R}^{d^r}$ of rank 1 for $r = 2$ and $r = 3$, respectively. Rank-1 tensors are then averaged by Avg resulting in tensor $\bar{\boldsymbol{\mathcal{V}}} \in \mathbb{R}^{d^r}$ that represents the contents of image $\mathcal{I}$ and could be used as a training sample. However, the practical image representations have to deal with so-called *burstiness* which is "*the property that a given visual element appears more times in an image than a statistically independent model would predict*" [11].

Corrective functions such as Power Normalization are known to suppress the burstiness [3, 25, 11]. Therefore, equations (17-19) and (20) apply two-stage pooling with eigenvalue- and coefficient-wise Power Normalization, respectively. In Equation (17), operator $\mathrm{HOSVD}: \mathbb{R}^{d^r} \to \left(\mathbb{R}^{d^r}; \mathbb{R}^{d \times d}\right)$ decomposes tensor $\bar{\boldsymbol{\mathcal{V}}}$ into a core tensor $\boldsymbol{E}$ of eigenvalues and an orthonormal factor matrix $\boldsymbol{A}$, which can be interpreted as the principal components in $r$ modes. Element-wise Power Normalization is then applied by equation (18) to eigenvalues $\boldsymbol{E}$ to even out their contributions. Tensor $\hat{\boldsymbol{\mathcal{V}}} \in \mathbb{R}^{d^r}$ is then assembled in equation (19) by the $r$-mode product $\times_r$ detailed in [13]. Lastly, coefficient-wise Power Normalization acting on $\hat{\boldsymbol{\mathcal{V}}}$ produces tensor $\boldsymbol{\mathcal{X}} \in \mathbb{R}^{d^r}$ in equation (20).

In our experiments, we use $r = 3$ and: i) $\boldsymbol{\mathcal{X}} = \bar{\boldsymbol{\mathcal{V}}}$ when Power Normalization is disabled, e.g., $\gamma_e = \gamma_c = 1$ or ii) apply the above HOSVD approach if $0 < \gamma_e < 1 \land 0 < \gamma_c \leq 1$. Therefore, the final descriptor $\boldsymbol{\mathcal{X}}$ in equation (20) represents an image by a super-symmetric third-order tensor which is guaranteed to contain SPD matrix slices if no Power Normalization is used.

## B  Derivation of the Third-order Aggregation.

Simple operations such as rising $K_{rbf}$ to the power of $r$ and applying outer product of order $r$ to $\boldsymbol{v}_{xy}$ form a higher-order tensor of rank-1 denoted as $\boldsymbol{\mathcal{V}}_{xy}$:

$$K_{rbf}^r((x, y, \mathbf{I}^a), (x', y', \mathbf{I}^b)) \approx \left\langle \boldsymbol{\mathcal{V}}_{xy}^a, \boldsymbol{\mathcal{V}}_{x'y'}^b \right\rangle, \text{where } \boldsymbol{\mathcal{V}}_{xy}^a = \otimes_r \boldsymbol{v}_{xy}^a. \tag{21}$$

The higher-order rank-1 tensors $\boldsymbol{\mathcal{V}}_{xy}$ are aggregated over image regions $\boldsymbol{\mathcal{R}}^a$ and $\boldsymbol{\mathcal{R}}^b$, respectively:

$$\sum_{(x,y) \in \boldsymbol{\mathcal{R}}^a} K_{rbf}^r((x, y, \mathbf{I}^a), (x', y', \mathbf{I}^b)) \approx \left\langle \bar{\boldsymbol{\mathcal{V}}}^a, \bar{\boldsymbol{\mathcal{V}}}^b \right\rangle, \bar{\boldsymbol{\mathcal{V}}}^a = \underset{(x,y) \in \boldsymbol{\mathcal{R}}^a}{\mathrm{Avg}} \boldsymbol{\mathcal{V}}_{xy}^a, \boldsymbol{\mathcal{V}}_{xy}^a = \otimes_r \boldsymbol{v}_{xy}^a, \tag{22}$$

$$\text{where } (x', y') \in \boldsymbol{\mathcal{R}}^b : x' = x - x_0^a + x_0^b \land y' = y - y_0^a + y_0^b. \tag{23}$$

The aggregation step in equation (22) provides analogy to both equation (16) and step (ii) which outlines the second-order aggregation step. Equation (23) highlights that the sum kernel is computed over locations $(x, y) \in \boldsymbol{\mathcal{R}}^a$ and corresponding to them locations $(x', y') \in \boldsymbol{\mathcal{R}}^b$. For instance, we subtract region origin $a_0^a$ and replace with origin $a_0^b$ to obtain $x'$ from $\boldsymbol{\mathcal{R}}^b$ that corresponds to $x$ from $\boldsymbol{\mathcal{R}}^a$. The higher-order autocorrelation tensors $\bar{\boldsymbol{\mathcal{V}}}$ are formed and can be then whitened by applying equations (17-20). In our experiments, we assume the choice of $r = 3$. Thus, the third-order dictionary learning and sparse coding is applied to form so-called mid-level features (one per region). Given an image and a set of overlapping regions, we apply pooling [14] over the mid-level features and obtain the final image signature that is used as a data sample to learn a classifier.